\documentclass[11pt,a4paper]{article}\pdfoutput=1

\usepackage{authblk}
\usepackage[hyperref]{naaclhlt2019}
\usepackage{times}
\usepackage{amsmath,amssymb,latexsym}
\usepackage{booktabs,multirow}
\usepackage{fontawesome}
\usepackage{adjustbox}
\usepackage{url}

\DeclareMathOperator{\tfidf}{\mathrm{tf}\!\textrm{--}\mathrm{idf}}

\aclfinalcopy

\title{HHMM at SemEval-2019 Task 2: Unsupervised Frame Induction using Contextualized Word Embeddings}

\author[$\star$]{\textbf{Saba Anwar}}
\author[$\dag$]{\textbf{Dmitry Ustalov}}
\author[$\ddag,\S$]{\textbf{Nikolay Arefyev}}
\author[$\dag$]{\textbf{Simone Paolo Ponzetto}}
\author[$\star$]{\\\textbf{Chris Biemann}}
\author[$\star$,$\diamond$]{\textbf{Alexander Panchenko}}

\affil[$\star$]{Language Technology Group, Department of Informatics, University of Hamburg, Germany}
\affil[$\diamond$]{Skolkovo Institute of Science and Technology, Russia}
\affil[ ]{\texttt{\{anwar,biemann,panchenko\}@informatik.uni-hamburg.de}}
\affil[$\dag$]{Data and Web Science Group, University of Mannheim, Germany}
\affil[ ]{\texttt{\{dmitry,simone\}@informatik.uni-mannheim.de}}
\affil[$\ddag$]{Samsung R\&D Institute Russia}
\affil[$\S$]{Lomonosov Moscow State University, Russia}
\affil[ ]{\texttt{narefyev@cs.msu.ru}}

\date{}

\begin{document}

\maketitle

\begin{abstract}
We present our system for semantic frame induction that showed the best performance in Subtask~B.1 and finished as the runner-up in Subtask~A of the SemEval~2019 Task 2 on unsupervised semantic frame induction~\citep{QasemiZadeh:19}. Our approach separates this task into two independent steps: verb clustering using word and their context embeddings and role labeling by combining these embeddings with syntactical features. A simple combination of these steps shows very competitive results and can be extended to process other datasets and languages.
\end{abstract}

\section{Introduction}

Recent years have seen a lot of interest in computational models of frame semantics, with the availability of annotated sources like PropBank~\citep{Palmer:05} and FrameNet~\citep{Baker:98}. Unfortunately, such annotated resources are very scarce due to their language and domain specificity. Consequently, there has been work that investigated methods for unsupervised frame acquisition and parsing~\citep{Lang:10,Modi:12,Kallmeyer:18,Ustalov:18:triframes}. Researchers have used different approaches to induce frames, including clustering verb-specific arguments as per their roles~\citep{Lang:10}, subject-verb-object triples~\cite{Ustalov:18:triframes}, syntactic dependency representation using dependency formats like CoNLL~\cite{Modi:12,Titov:12}, and latent-variable PCFG models~\cite{Kallmeyer:18}.

The SemEval~2019 task of semantic frame and role induction consists of three subtasks: (A) learning the frame type of the highlighted verb from the context in which it has been used; (B.1) clustering the highlighted arguments of the verb into specific roles as per the frame type of that verb, e.g., \textit{Buyer}, \textit{Goods}, etc.; (B.2) clustering the arguments into generic roles as per VerbNet classes~\citep{Schuler:05}, without considering the frame type of the verb, i.e., \textit{Agent}, \textit{Theme}, etc.

Our approach to frame induction is similar to the word sense induction approach by \citet{Arefyev:18}, which uses $\tfidf$-weighted context word embeddings for a shared task on word sense induction by \citet{Panchenko:18:russe}. In this unsupervised task, our approach for clustering mainly consists of exploring the effectiveness of already available pre-trained models.\footnote{HHMM is an abbreviation for Hansestadt Hamburg, Mannheim, and Moscow. It is chosen to avoid confusion with hidden Markov models.} Main contributions of this paper are:
\begin{enumerate}
  \item a method that uses contextualized distributional word representations (embeddings) for grouping verbs to frame type clusters (Subtask~A);
  \item a method that combines word and context embeddings for clustering arguments of verbs to frame slots (Subtasks~B.1 and B.2).
\end{enumerate}

The key difference of our approach with respect to  prior work by \citet{Arefyev:18} and \citet{Kallmeyer:18} is that we have only used pre-trained embeddings to disambiguate the verb senses and then combined these embeddings with additional features for semantic labeling of the verb roles.\footnote{Our code is available at \url{https://github.com/uhh-lt/semeval2019-hhmm}.}

The remainder of the paper is organized as follows. The methodology and the results for each subtask are discussed in Sections~\ref{sec:task1}, \ref{sec:task21} and \ref{sec:task22} respectively, followed by the conclusion in Section~\ref{sec:conclusion}.

\section{\label{sec:task1}Subtask~A: Grouping Verbs to Frame Type Clusters}

In this subtask, each sentence has a highlighted verb, which is usually the predicate. The goal is to label each highlighted verb according to the frame evoked by the sentence. The gold standard for this subtask is based on the FrameNet~\citep{Baker:98} definitions for frames.

\subsection{Method}

Since sentences evoking the same frame should receive the same labels, we used a verb clustering approach and experimented with a number of pre-trained word and sentence embeddings models, namely Word2Vec~\citep{Mikolov:13}, ELMo~\citep{Peters:18}, Universal Sentence Embeddings~\citep{Conneau:17}, and fastText~\citep{Bojanowski:17}. This setup is similar to treating the frame induction task as a word sense disambiguation task~\citep{Brown:11}.

We experimented with embedding different lexical units, such as verb (\textsc{v}), its sentence (context, \textsc{c}), subject-verb-object (SVO) triples, and verb arguments. Combination of context and word representations (\textsc{c}+\textsc{w}) from Word2Vec and ELMo turned out to be the best combination in our case.

We used the standard Google News Word2Vec embedding model by \citet{Mikolov:13}. Since this model is trained on individual words only and the SemEval dataset contained phrasal verbs, such as \textit{fall back} and \textit{buy out}, we have considered only the first word in the phrase. If this word is not present in the model vocabulary, we fall back to a zero-filled vector. When aggregating a context into a vector, we used the $\tfidf$-weighted average of the word embeddings for this context as proposed by \citet{Arefyev:18}. We tuned these weights on the development dataset.

We used the ELMo contextualized embedding model by \citet{Peters:18} that generates vectors of a whole context. Similarly to fastText~\citep{Bojanowski:17}, ELMo can produce character-level word representations to handle out-of-vocabulary words. In all our experiments we used the same pre-trained ELMo model available on TensorFlow Hub.\footnote{\url{https://tfhub.dev/google/elmo/2}} Among all the layers of this model, we used the mean-pooling layer for word and context embeddings.

\subsection{Results and Discussion}

\begin{table}[t]
\centering\small
\begin{tabular}{cp{40mm}rr}\toprule
\multicolumn{2}{l}{\textbf{Method}} & \textbf{Pu\ F\textsubscript{1}} & \textbf{B\textsuperscript{3}\ F\textsubscript{1}} \\\midrule
{\faFire} & w2v[\textsc{c}+\textsc{w}]norm & $\mathbf{76.68}$ & $\mathbf{68.10}$ \\
{\faHourglassEnd} & ELMo[\textsc{c}+\textsc{w}]norm & $77.03$ & $69.50$ \\\midrule
{\faTachometer} & Cluster Per Verb & $73.78$ & $65.35$ \\
{\faTrophy} & Winner & $78.15$ & $70.70$ \\\bottomrule
\end{tabular}
\caption{\label{tab:task1}Our results on Subtask~A: Grouping Verbs to Frame Type Clusters. Purity F\textsubscript{1}-score is denoted as \textit{Pu\ F\textsubscript{1}}, B-Cubed F\textsubscript{1}-score is denoted as \textit{B\textsuperscript{3}~F\textsubscript{1}}. {\faFire} denotes \textit{our} final submission (\#~536426), {\faHourglassEnd} denotes \textit{our} post-competition result, {\faTachometer} denotes a baseline, and {\faTrophy} denotes the submission of the winning team.}
\bigskip
\end{table}

We experimented with different clustering algorithms provided by scikit-learn~\citep{Pedregosa:11}, namely agglomerative clustering, DBSCAN, and affinity propagation. After the model selection on the development dataset, we have chosen agglomerative clustering for further evaluation. Although both ELMo and Word2Vec showed the best results on the development dataset with single linkage, we opted average linkage after analyzing $t$-SNE plots~\citep{vanderMaarten:08}.

\tablename~\ref{tab:task1} shows our results obtained on Subtask~A. Our final submission ({\faFire}) used agglomerative clustering of normalized vectors obtained by concatenating the context and verb vectors from the Word2Vec model. In particular, we found that the best performance is attained for Manhattan affinity and 150 clusters. During our post-competition experiments ({\faHourglassEnd}), we found that ELMo performed better than Word2Vec when a higher number of clusters, 235, was specified.

\section{\label{sec:task21}Subtask~B.1: Clustering Arguments of Verbs to Frame-Specific Slots}

In this subtask, each sentence has a set of highlighted nouns or noun phrases corresponding to the slots of the evoked frame. Additionally, each sentence is provided with the same highlighted verb as in Subtask~A (Section~\ref{sec:task1}). The goal is to label each highlighted verb according to the evoked frame and to assign each highlighted token a frame-specific semantic role identifier. The gold standard for this subtask is annotated with FrameNet frames and roles~\citep{Baker:98}.

\subsection{Method}

Since Subtask~B.1 asks to assign role labels to highlighted tokens as per the frame type of the verb, we attempted this by merging the output of verb frame types from Subtask~A (Section \ref{sec:task1}) and the output of generic role labels from Subtask~B.2 (Section \ref{sec:task22}). We used \texttt{UKN} (unknown) slot identifier for the tokens present in Subtask~B.1, but missing in Subtask~B.2.

\subsection{Results and Discussion}

\begin{table}[t]
\centering
\resizebox{1\linewidth}{!}{
\begin{tabular}{cp{70mm}rr}\toprule
\multicolumn{2}{l}{\textbf{Method}} & \textbf{Pu\ F\textsubscript{1}} & \textbf{B\textsuperscript{3}\ F\textsubscript{1}} \\\midrule
\multicolumn{4}{l}{\textit{Agglomerative Clustering}} \\
\multirow{2}{*}{\faFire} & Subtask~A: w2v[\textsc{c}+\textsc{w}] & \multirow{2}{*}{$\mathbf{62.10}$} & \multirow{2}{*}{$\mathbf{49.49}$} \\
& Subtask~B.2: ID & \\\midrule
\multicolumn{4}{l}{\textit{Logistic Regression}} \\
\multirow{2}{*}{\faFlash} & Subtask~A: w2v[\textsc{c}+\textsc{w}]norm & \multirow{2}{*}{$66.81$} & \multirow{2}{*}{$55.61$} \\
& Subtask~B.2: ELMo[\textsc{c}+\textsc{w}+\textsc{v}]+ID+B+123 & \\
\multirow{2}{*}{\faHourglassEnd} & Subtask~A: ELMo[\textsc{c}+\textsc{w}]norm & \multirow{2}{*}{$68.22$} & \multirow{2}{*}{$58.61$} \\
& Subtask~B.2: w2v[\textsc{c}+\textsc{w}+\textsc{v}]+ID+B+123 & \\\midrule
{\faTachometer} & Cluster Per Dependency Role & $57.99$ & $45.79$ \\
{\faTrophy} & Winner & $62.10$ & $49.49$ \\\bottomrule
\end{tabular}
}
\caption{\label{tab:task21}Our results on Subtask~B.1: Clustering Arguments of Verbs to Frame-Specific Slots. Purity F\textsubscript{1}-score is denoted as \textit{Pu\ F\textsubscript{1}}, B-Cubed F\textsubscript{1}-score is denoted as \textit{B\textsuperscript{3}~F\textsubscript{1}}. {\faFire} denotes \textit{our} final submission (\#~535483), {\faFlash} denotes a supervised \textit{Logistic Regression} submission that does not comply to the task rules, {\faHourglassEnd} denotes \textit{our} post-competition result, {\faTachometer} denotes a baseline, and {\faTrophy} denotes the submission of the winning team.}
\end{table}

\tablename~\ref{tab:task21} shows the results from merging our solutions for Subtasks~A and B.2, as described in Sections~\ref{sec:task1} and \ref{sec:task22}, correspondingly. For our final submission ({\faFire}), we merged the frame types obtained by clustering the Word2Vec embeddings of the sentence (context, \textsc{c}) and verb (word, \textsc{w}), and the role labels obtained by clustering the vector of inbound dependencies (ID). However, we observed that the logistic regression model demonstrated better performance in Subtask~B.2 than any clustering technique we tried, including our final submission~({\faFire}) and the baselines. But this performance was further improved by combining the results from post-competition experiments of Subtask~A and Subtask~B.2~({\faHourglassEnd}).

\section{\label{sec:task22}Subtask~B.2: Clustering Arguments of Verbs to Generic Roles}

In Subtask~B.2, similarly to Subtask~B.1 (Section~\ref{sec:task21}), each sentence has a set of highlighted nouns or noun phrases that correspond to the slots of the evoked frame. The goal is to label each highlighted token with a high-level generic class, such as \textit{Agent} or \textit{Patient}. However, unlike Subtask~B.1, the verb frame labeling part is omitted. The gold standard for this subtask is annotated as according to the VerbNet classes~\citep{Schuler:05}.

\subsection{Method}

When addressing this subtask, we experimented with combining the embeddings of the word (\textsc{w}) filling the role, its sentence (context, \textsc{c}), and the highlighted verb (\textsc{v}). To handle the out-of-vocabulary roles in the case of Word2Vec embeddings, each role was tokenized and embeddings for each token were averaged. If a token is still not present in the vocabulary, then a zero-filled vector was used as its embedding. During prototyping we developed several features that improved the performance score, namely \textit{inbound dependencies} (ID), which represent the dependency label from the head to the role (dependent) and two trivial baselines: \textit{Boolean} (B) and \textit{123}.

We built a negative one-hot encoding feature vector to represent the \textit{inbound} dependencies of the word corresponding to the role. Thus, for each dependency of the given role~(in case of a multi-word expression), we fill \texttt{-1} if the dependency relationship holds, otherwise \texttt{0} is filled. During our experiments for the development test, we also used the outbound dependencies, which represent the dependency label from the role (head) to the dependent words. So we used \texttt{-1} for inbound and \texttt{1} for outbound. But since they did not perform well in comparison to inbound dependencies, they were not considered for submitted runs.

For the \textit{Boolean} baseline, given the position of the verb in the sentence $p_v$ and the position of the target token $p_t$, we assign the role~\texttt{0} to $t$ if ${p_v < p_t}$, otherwise ~\texttt{1}. For the \textit{123} baseline, we assign its index to each highlighted slot filler. For example, if five slots need to be labelled, the first one will be labelled as \texttt{1} and the last one will be labelled as~\texttt{5}.

\subsection{Results and Discussion}

\begin{table}[t]
\centering
\resizebox{1.0\linewidth}{!}{
\begin{tabular}{cp{50mm}rr}\toprule
\multicolumn{2}{l}{\textbf{Method}} & \textbf{Pu\ F\textsubscript{1}} & \textbf{B\textsuperscript{3}\ F\textsubscript{1}} \\\midrule
\multicolumn{4}{l}{\textit{Agglomerative Clustering}} \\
{\faFire} & w2v[\textsc{c}]+ID & $\mathbf{62.00}$ & $\mathbf{42.10}$ \\
{\faHourglassEnd} & ELMo[\textsc{c}]+ID & $50.37$ & $34.89$ \\\midrule
\multicolumn{4}{l}{\textit{Logistic Regression}} \\
{\faFlash} & ELMo[\textsc{c}+\textsc{w}+\textsc{v}]+ID+B+123 & $73.14$ & $57.37$ \\
{\faHourglassEnd} & w2v[\textsc{c}+\textsc{w}+\textsc{v}]+ID+B+123 & $74.36$ & $58.83$ \\\midrule
\multirow{3}{*}{\faTachometer} & Cluster Per Dependency Role & $56.05$ & $39.03$ \\
& Boolean Baseline & $67.16$ & $46.78$ \\
& Inbound Dependencies (ID) & $66.05$ & $45.77$ \\\midrule
{\faTrophy} & Winner & $64.16$ & $45.65$ \\\bottomrule
\end{tabular}
}
\caption{\label{tab:task22}Our results on Subtask~B.2: Clustering Arguments of Verbs to Generic Roles. Purity F\textsubscript{1}-score is denoted as \textit{Pu\ F\textsubscript{1}}, B-Cubed F\textsubscript{1}-score is denoted as \textit{B\textsuperscript{3}~F\textsubscript{1}}. {\faFire} denotes \textit{our} final submission (\#~535480), {\faFlash} denotes a supervised \textit{Logistic Regression} submission that does not comply to the task rules, {\faHourglassEnd} denotes \textit{our} post-competition result, {\faTachometer} denotes a baseline, and {\faTrophy} denotes the submission of the winning team.}
\end{table}

\tablename~\ref{tab:task22} shows our results on Subtask~B.2. We found that the trivial Boolean approach outperformed LPCFG~\citep{Kallmeyer:18} and all the standard baselines, including cluster per dependency role (\texttt{OneClustPerGrType}), on the development dataset.\footnote{On the development dataset for Subtask~B.2, the \textit{Boolean} baseline demonstrated B-Cubed $F_1=57.98$, while LPCFG and cluster per dependency role yielded $F_1=40.05$ and ${F_1=50.79}$, correspondingly.}

Similarly to our solution for Subtask~A (Section~\ref{sec:task1}), we tried different clustering algorithms to cluster arguments of verbs to generic roles and found that the best clustering performance is shown by agglomerative clustering with Euclidean affinity, Ward's method linkage, and two clusters. Our final submission ({\faFire}) used the combination of inbound dependencies and Word2Vec embedding for sentence (context, \textsc{c}), which performed marginally better than the cluster per dependency role (\texttt{OneClustPerGrType}) baseline, but still not better than such trivial baselines as \textit{Boolean} or $123$. Replacing Word2Vec with ELMo in our post-competition experiments have lowered the performance further.

In order to estimate our upper bound of the performance, we compared our best-performing clustering algorithm, i.e., agglomerative clustering, to a logistic regression model. We found that the combination of sentence (context, \textsc{c}), target word (\textsc{w}), and verb (\textsc{v}) vectors, enhanced with our other features, shows substantially better results than a simple clustering model ({\faFlash}). However, we did not observe a noticeable difference between the performance of the underlying embedding models.

As the model was trained on the development dataset that contained $20$ roles in contrast to the test set which contained $32$ roles, this approach has its limitations due to this difference of the number and meaning of roles. We believe that the performance could be improved using semi-supervised clustering methods, yet during prototyping with the pairwise-constrained $k$-Means algorithm~\citep{Basu:04} we did not observe any performance improvements.

\section{\label{sec:conclusion}Conclusion}

We presented an approach for unsupervised semantic frame and role induction that uses word and context embeddings. It separates the task into two independent steps: verb clustering and role labelling, using combination of these embeddings enhanced with syntactical features. Our approach showed the best performance in Subtask~B.1 and also finished as the runner-up in Subtask~A of this shared task, and it can be easily extended to process other datasets and languages.

\subsubsection*{Acknowledgments}

We acknowledge the support of the Deutsche Forschungsgemeinschaft (DFG) under the ``JOIN-T'' and ``ACQuA'' projects and the German Academic Exchange Service (DAAD). We thank the SemEval organizers for an inspiring shared task and their quick responses to all our questions. We are grateful to four anonymous reviewers who offered useful comments. Finally, we thank the Linguistic Data Consortium (LDC) for the provided Penn Treebank dataset~\citep{Marcus:93}.


\bibliographystyle{acl_natbib}
\bibliography{hhmm-semeval2019-task2}

\end{document}